\documentclass[letterpaper, 10 pt, conference]{ieeeconf}  

\IEEEoverridecommandlockouts                              

\newcommand{\mname}{\textsl{NavigateDiff}}
\usepackage{booktabs}
\usepackage{graphicx}
\usepackage{multirow}
\usepackage{algorithm}
\usepackage{algorithmic}
\usepackage{amsmath}
\usepackage{mathrsfs} 
\usepackage{amssymb}
\usepackage{subfigure}
\usepackage[colorlinks=false]{hyperref}

\title{\LARGE \bf
NavigateDiff: Visual Predictors are Zero-Shot Navigation Assistants
}

\author{Yiran Qin$^{1,2*}$, Ao Sun$^{2*}$, Yuze Hong$^{2}$, Benyou Wang$^{2}$, Ruimao Zhang$^{1\dag}$ \\
\normalsize$^1$ Sun Yat-sen University \qquad 
\normalsize$^2$ The Chinese University of Hong Kong, Shenzhen \\
\tt\small yiranqin@link.cuhk.edu.cn~~~aosun3@link.cuhk.edu.cn~~~ruimao.zhang@ieee.org\\
\thanks{$*$ Equal Contribution; $^{\dag}$ Corresponding Author}
\thanks{The work is partially supported by Shenzhen Science and Technology Program JCYJ20220818103001002, the Guang.dong Key Laboratory of Big Data Analysis and Processing, Sun Yat-sen University, China, and by the High-performance Computing Public Platform (Shenzhen Campus) of Sun Yat-sen University.}
}

\begin{document}

\maketitle
\thispagestyle{empty}
\pagestyle{empty}

\begin{abstract}

Navigating unfamiliar environments presents significant challenges for household robots, requiring the ability to recognize and reason about novel decoration and layout. 
Existing reinforcement learning methods cannot be directly transferred to new environments, as they typically rely on extensive mapping and exploration, leading to time-consuming and inefficient.
To address these challenges, we try to transfer the logical knowledge and the generalization ability of pre-trained foundation models to zero-shot navigation. 
By integrating a large vision-language model with a diffusion network, our approach named \mname ~constructs a visual predictor that continuously predicts the agent's potential observations in the next step which can assist robots generate robust actions.
Furthermore, to adapt the temporal property of navigation, we introduce temporal historical information to ensure that the predicted image is aligned with the navigation scene.
We then carefully designed an information fusion framework that embeds the predicted future frames as guidance into goal-reaching policy to solve downstream image navigation tasks.
This approach enhances navigation control and generalization across both simulated and real-world environments. 
Through extensive experimentation, we demonstrate the robustness and versatility of our method, showcasing its potential to improve the efficiency and effectiveness of robotic navigation in diverse settings.
Project Page: \href{https://21styouth.github.io/NavigateDiff/}{https://21styouth.github.io/NavigateDiff/}.

\end{abstract}

\section{Introduction}

A useful generalist navigation robot must be able to, much like a human, recognize and reason about novel environments and pathways it has never encountered before. 
For example, if a user instructs the robot to ``find and navigate to that red building," the robot should be able to accomplish this task even if it has never navigated in that environment or even seen a red building before. 
In other words, the robot needs not only the physical capability to move through complex environments but also the ability to understand the underlying world rules and perform logical reasoning in environments beyond its training distribution.
Although the size of datasets used for robotic navigation has increased in recent years, these datasets cannot possibly cover every environment and scenario the robot might encounter—just as a person's life experiences cannot include physical interactions with every type of environment. 
While these datasets provide numerous examples of navigation scenarios, they lack the broad logical knowledge required for robots to navigate the specific environments they may encounter in everyday tasks.

\begin{figure}[thpb]
  \centering 
  \includegraphics[width=0.45\textwidth]{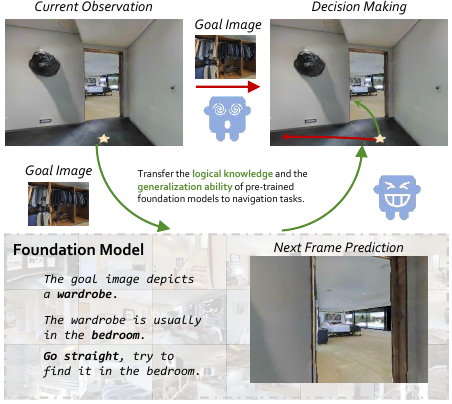}
  \caption{\mname ~leverages logical reasoning and generalization capabilities of pre-trained foundation models to enhance zero-shot navigation by predicting future observations to assist robot generate robust actions}
  \label{fig:motivation}
\end{figure}

Given this, one question naturally arises: how can we integrate this logical knowledge into navigation control? 
Recently, machine learning methods have achieved broad success in natural language processing~\cite{brown2020language}, visual perception~\cite{podell2023sdxl,yu2025gamefactory,qin2024worldsimbench,li2025t2isafety,an2024agfsync}, and other domains~\cite{brohan2022rt,brohan2023rt,huang2024story3d,zhang2024ad} by leveraging Internet-scale data to train general-purpose “foundation” models. 
These models are equipped with extensive logical knowledge and can adapt to new tasks through zero-shot transfer, prompt tuning, or fine-tuning on target data~\cite{qin2023mp5,li2024manipllm,zhou2024code}. 
Building on these advancements, researchers in the navigation field have begun to utilize pre-trained foundation models in vision and language to build navigation systems~\cite{shah2023vint} or use pre-trained vision-language encoders to initialize navigation policies~\cite{ lin2024navcot, NavGPT}.
Although these methods aim to incorporate the logical knowledge in Internet-scale data into navigation model training, they (1) often fail to fully exploit the diversity of Internet data, instead being confined to navigation-specific data; (2) merely improve high-level text instruction generalization, which is difficult to apply to specific navigation tasks as the policy network cannot grasp abstract textual concepts.

%

To address this, we propose \mname, a novel navigation framework to realize generalizable navigation, which introduces the visual predictors to bridge the gap between the high-level trajectory planning in the scene and low-level policy generation for the robot control.
As illustrated in Fig.~\ref{fig:motivation}, unlike existing approaches that directly generate robotic policies using current observations and goal images, the proposed method leverages the logical knowledge of foundation models to infer the robot's next state and generate the corresponding visual content. The results of this visual prediction, along with the original goal image, are then used to assist the policy network in generating specific actions for the robot.
In practice, We first construct a \textbf{Predictor} by leveraging a large vision-language model followed by a diffusion model. This predictor reasons about the next predicted frame, which is helpful for navigation task execution, by leveraging the goal image, current observation, and optional additional information (such as text instructions or historical observation data). 
Next, to achieve robot control, we also train a \textbf{Hybrid Fusion Policy Network}, which outputs the final control signal by integrating multiple visual information sources, \textit{i.e.} the goal image, current observation, and generated future frame.

%
%

Our proposed method has achieved impressive navigation results in both simulation and real-world scenarios. These results are attributed to the fact that our framework decouples the complex high-level reasoning and efficient low-level control required for navigation in open environments.
Specifically, when the model needs to analyze and reason about navigation tasks, it can directly leverage the logical knowledge of the vision-language model for effective reasoning, bypassing the need to understand the low-level dynamics of the robot.
Conversely, when the model is required to control the robot, the policy network can focus solely on the relationship between the current state and the immediate next state (\textit{i.e.} generated future frame), without the burden of considering long-term task logic or intricate environmental rules.
%
%
%
%
%
%
In summary, the main contributions are as follows:
\begin{itemize}
%
\item We propose \mname, a novel navigation framework designed to separate high-level task reasoning from low-level robot control. This decoupling enhances the effectiveness and generalizability of visual navigation.

\item We introduce a Predictor that harnesses the advanced reasoning capabilities of large vision-language models to produce a plausible intermediate frame, facilitating downstream robotic control. Additionally, the Hybrid Fusion Policy Network ensures that the generated actions are more stable in practice.
%
%
\item We conduct extensive experiments in both simulated and real-world settings, yielding numerous experimental results that validate the effectiveness and robustness of our proposed framework.
\end{itemize}
\section{Related Work}

\subsection{Vision-based Navigation}


Classical SLAM-based methods~\cite{chaplot2020neural} and learning-based approaches~\cite{maksymets2021thda,mezghan2022memory,wang2024toward,qin2023supfusion} have both been applied to embodied visual navigation. While end-to-end learning techniques tend to rely less on manually designed components, they have demonstrated greater potential. For instance, Memory-Augmented RL~\cite{mezghan2022memory} employs an attention mechanism that uses episodic memory to facilitate navigation, achieving state-of-the-art results in ImageNav~\cite{zhu2017target} with four RGB cameras. In contrast, our model adopts a more straightforward architecture and still achieves superior performance. For single-camera setups, \cite{al2022zero} enhances performance by incorporating both goal-view rewards and goal-view sampling. We observe that applying this reward system and view sampling leads to further gains for OVRL models.

Similarly, end-to-end reinforcement learning methods have been applied to ObjectNav, utilizing data augmentation~\cite{maksymets2021thda} and auxiliary rewards~\cite{ye2021auxiliary} to enhance generalization. On the other hand, modular methods~\cite{chaplot2020object,ramakrishnan2022poni} separate the tasks of navigation and semantic mapping. A recent competitive imitation learning approach~\cite{ramrakhya2022habitat}, built on a large-scale dataset, offers another direction, which our work builds upon. In contrast to~\cite{mousavian2019visual}, which focuses on improving visual representations by incorporating semantic segmentation, we prioritize RGB-based representations in our approach.

\subsection{Diffusion Models for Image Generation}


Recent advances in text-to-image diffusion models~\cite{dhariwal2021diffusion, ho2022classifier,nichol2021glide} have greatly improved instruction-driven image-to-image methods~\cite{brooks2023instructpix2pix,fu2023guiding,geng2024instructdiffusion,zhou2024minedreamer} like InstructPix2Pix~\cite{brooks2023instructpix2pix}, primarily used for image editing that aims to alter content while keeping the background constant. 
The method struggles with egocentric images in dynamic, navigation tasks, where the backgrounds change significantly and the images should conform more to physical rules.
In this work, we train the MLLM-enhanced diffusion model to generate continuous future images for guiding real-time, low-level control for image navigation tasks.

\subsection{Pretrained Foundation Models for Embodied Tasks}


Foundation models like Large Language Models (LLMs) and Diffusion Models have become a powerful tool for prior-knowledge reasoning in navigation, due to their information processing and generative abilities. For example, LLMs are used to predict correlations with the target object at both the object and room levels, aiding in locating the target~\cite{zhou2023esc, gadre2023cows}. They also help cluster unexplored areas and infer relationships between the target and surrounding objects to guide navigation~\cite{yu2023l3mvn, cai2024bridging}. Chain-of-thought (CoT) reasoning is integrated into LLMs to encourage exploration in areas likely to contain the target while avoiding irrelevant regions~\cite{shah2023navigation}. In multi-robot collaborative navigation, LLMs centrally plan mid-term goals by analyzing data such as obstacles, frontiers, and object coordinates from online maps~\cite{yu2023co}. Furthermore, path and farsight descriptions are combined to enable LLMs to apply commonsense reasoning for waypoint identification~\cite{wu2024voronav}.
Additionally, Diffusion Models have made significant advances in embodied scenarios. For instance, video diffusion is combined with inverse dynamics to generate robot control signals for specific tasks~\cite{du2024learning, ajay2024compositional}. In another approach, diffusion models are used for interpretable hierarchical planning through skill abstractions during task execution~\cite{liang2024skilldiffuser}. Furthermore, these models are employed to guide agents through open-ended tasks~\cite{qin2023mp5, zhou2024minedreamer}.
\begin{figure*}[ht]
    \centering
    \includegraphics[width=0.99\textwidth]{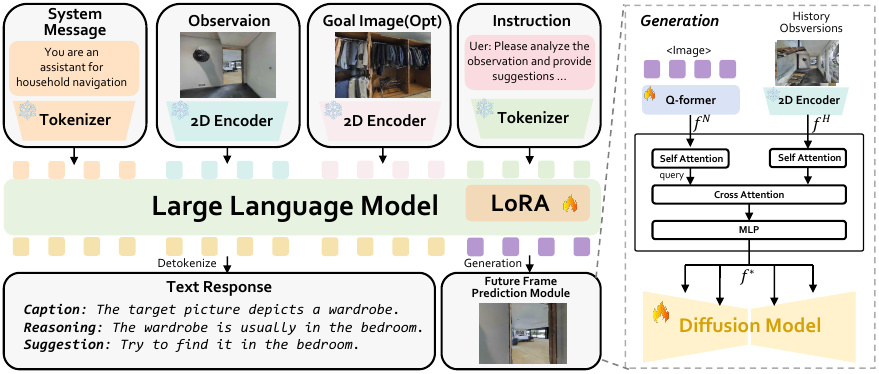}
    \caption{\textbf{The Overall Architecture of Predictor.} Instruction, current observation, and goal image are encoded separately and sent into LLaVA~\cite{liu2024visual}. Then LLaVA~\cite{liu2024visual} generates hidden states for the Special Image Tokens $<$image$>$ tokens, we transform $<$image$>$ into semantically relevant representations $f^N$ using a Q-Former. The feature $f^H$, extracted from the 2D encoder, is fused with the feature $f^N$. The resulting fused feature $f^*$ is then used as a condition in the Edit-based diffusion model to generate future images.}
    \label{fig:generator}
\end{figure*}

\section{Method}

\mname ~utilizes the logical knowledge and the generalization ability of pre-trained foundation models to improve zero-shot navigation in the presence of novel environments, scenes, and objects. 
How can we achieve this when foundation models trained on general-purpose Internet data do not provide direct guidance for selecting low-level navigation actions? 
Our key insight is to decouple the navigation problem into two stages: (I) generating intermediate future goals that need to be reached to successfully navigate, and (II) learning low-level control policies for reaching these future goals. 
In Stage (I), we build a \textbf{Predictor} by incorporating a Multimodal Large Language Model (MLLM) with a diffusion model, fine-tuned with parameter efficiency, specifically designed for Image Navigation~\cite{zhu2017target}. 
Stage (II) involves training a \textbf{Fusion Navigation Policy} on image navigation data and testing in new environments. We describe data collection and each of these stages in detail below and summarize the resulting navigation algorithm.

\subsection{Data Formulation}
\label{sec_data}
For the generation of future frame training data, we utilized the simulator's built-in ``shortest path follower" algorithm in the simulation environment to obtain the standard route for each task and generate corresponding videos. 
In the real world, we recorded ego-view perspective videos of a human remotely controlling a navigation robot to complete the image navigation tasks.
Based on the collected videos, we randomly selected a starting frame from each video and chose the corresponding future frame according to a predefined prediction interval $k$. 
We also recorded the relevant navigation task information, ultimately forming the training tuples $(x_t, x_{t+k}, x_h, y, x_g)$.
where $x_t$ represents the current observation image, $x_{t+k}$ is the future frame image that needs to be predicted, $x_h$ is the history frame of $x_t$, $y$ indicates the task's corresponding textual instruction and $x_g$ is the final goal image for the navigation task.

\begin{figure*}[t]
    \centering
    \includegraphics[width=0.9\textwidth]{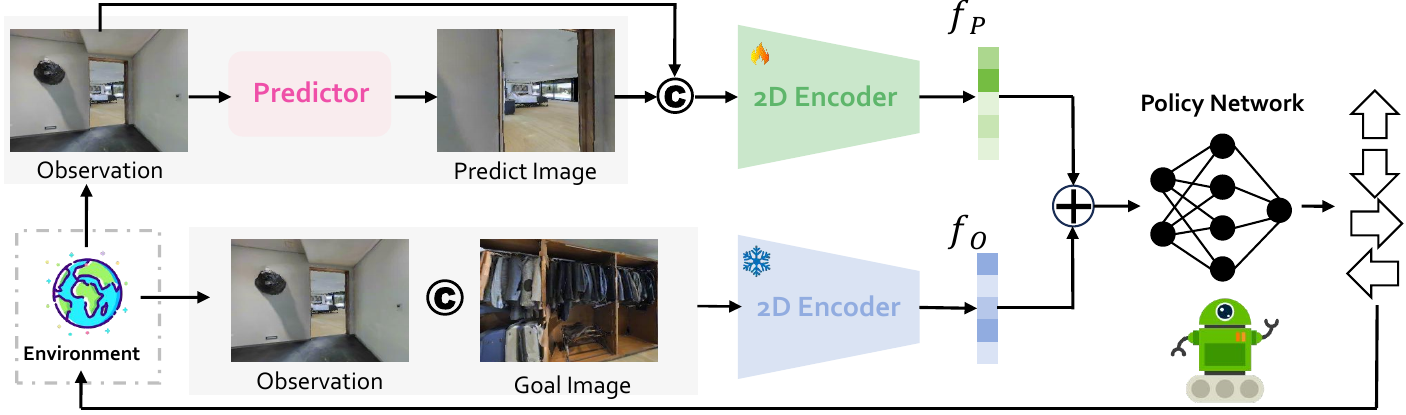}
    \caption{\textbf{\mname} ~leverages a Future Frame pretrained Predictor to generate future frames based on current observation and navigation task information. A fusion navigation policy then executes the actions needed to reach the future frames. Alternating this loop enables us to accomplish the navigation task.}
    \label{fig:pipeline}
\end{figure*}

\subsection{Predictor}

As illustrated in Fig.~\ref{fig:generator}, our Predictor integrates a \textbf{Multimodal Large Language Model~(MLLM)} with a \textbf{Future Frame Prediction Model}, which could process current observations, target images, and instructions and generate a predicted future image based on the provided input.

 \textbf{Multimodal Large Language Model.} 
Given a current observation $x_t$, a goal image $x_g$, and a textual instruction $y$, the Predictor generates a future frame. 
As shown in Fig.~\ref{fig:generator}, the current observation and goal image are processed through a frozen image encoder, while the textual instruction is tokenized and passed to the LLM. 
The Predictor can now generate Special Image Tokens $<$image$>$ based on the instruction's intent, though initially limited to the language modality. 
These tokens are then sent to the Future Frame Prediction Model to produce the final future frame prediction.
In practice, we utilized LLaVA as our base model, which consists of a pre-trained CLIP visual encoder (ViT-L/14) and Vicuna-7B as the LLM backbone. 
To fine-tune the LLM, we applied LoRA, initializing a trainable matrix to adapt the model for the task.

 \textbf{Future Frame Prediction Model.} For future frame generation, we bridge the gap between the LLM's hidden states and the LLaVA text encoder's spaces, we transform the Special Image Tokens $<$image$>$ into semantically relevant representations $f^N$ using a Q-Former. 
The feature $f^H$, extracted from the 2D encoder, is fused with the feature $f^N$. 
The resulting fused feature $f^*$ is then used as a condition in the Edits-based diffusion model~\cite{koh2024generating} to generate future images:
{
\vspace{-2mm}
\begin{equation}
f^*=\mathbf{H}(\mathbf{Q}\left(h_{<image>}\right), \mathbf{E}_{v}(x_{h}))
\end{equation}
}
where $\mathbf{Q}$ denotes the Q-Former, $\mathbf{E}_{v}$ is a 2D encoder, $\mathbf{E}_{v}(x_{h})$ indicates history observations, and $\mathbf{H}$ is the fusion block that includes two self-attention blocks, a cross-attention block, and an MLP layer.

We employ a latent diffusion model with a VAE to perform denoising diffusion in the latent space, conditioned on fused feature $f^*$. Formally, the training objective is given by:
{
\begin{equation}
\begin{split}
\mathcal{L}_{\mathrm{predictor}}=\mathbb{E}_{\mathcal{E}(x_{t+k}), \mathcal{E}(x_{t}), \epsilon \sim \mathcal{N}(0,1), s}[\| \epsilon \nonumber -\\\epsilon_\delta(s, [z_s, \mathcal{E}(x_{t})]+f^*) \|_2^2]
\label{eq:diffusion}
\end{split}
\tag{2}
\end{equation}
}
where $\epsilon$ represents unscaled noise, $s$ denotes the sampling step, and $z_{s}$ is the latent noise at step $s$. The term $\mathcal{E}(x_t)$ corresponds to the current observation condition. 

By incorporating temporal information, this hybrid approach captures object trajectories and richer scene semantics, allowing for more accurate predictions of object movements and environmental changes. This makes our method particularly effective for zero-shot navigation in dynamic scenes. Dataset construction and implementation details can be found in \ref{exp_ffp}.

\subsection{Fusion Navigation Policy}

Although the Predictor provides one-step future state planning within the vision modality, we also need to train a low-level controller to select appropriate navigation actions. 
As the Predictor generates future frame images infused with prior knowledge from the pre-trained foundation model, our approach requires only the design of an image fusion navigation policy to effectively leverage the foundational model's logical reasoning and generalization capabilities for robot navigation tasks.

 \textbf{Image Fusion Policy.}
During the training phase, the current observation $x_t$ is concatenated with the future frame $x_{t+k}$ along the RGB channel dimension and processed through a trainable 2D Encoder to obtain $f_p$. The current observation $x_t$ is also concatenated with the final goal image $x_g$ along the RGB channel dimension and passed through a pre-trained 2D Encoder to obtain $f_o$. Based on the fused representations of the future and goal images, we train a navigation policy $\pi$ using reinforcement learning (RL):
\begin{equation}
s_t = \pi(~[~f_p, f_o, a_{t-1}~]~|~h_{t-1}~)
\end{equation}
where $s_t$ represents the embedding of the agent's current state, and $h_{t-1}$ denotes the hidden state of the recurrent layers in the policy $\pi$ from the previous step. We employ an actor-critic network~\cite{al2022zero} to predict both the state value and the action $a_t$ using $s_t$, training the model end-to-end with PPO \cite{schulman2017proximal}.

During the testing phase, we utilize trained Predictor and Fusion Navigation Policy to navigate in new environments based on corresponding goal images. For the image navigation task with a goal image $x_g$, given a current observation $x_t$, we generate the next future frame $x_t^* \leftarrow \mathcal{G}(x_t,x_g,y)$, once the future frame is sampled, we then execute the Fusion Navigation Policy $\pi$ conditioned on $x_t^*,x_t,x_g$ for $k$ timesteps, where $k$ is a testing hyperparameter. After $k$ timesteps, the future frame is refreshed by sampling from the Predictor again, and the process is repeated. Conventional wisdom suggests that regenerating future frames more frequently could result in more robust navigation control, assuming an unlimited computational budget. However, in practice, to maintain computational efficiency, we set $k$ to closely match the corresponding timesteps used during training, which has consistently delivered satisfactory performance. The pseudocode for navigation testing is detailed in Algorithm.~\ref{alg:NavigateDiff}.

\begin{algorithm}[ht]
\small
\caption{NavigateDiff: Zero-Shot Navigation Testing}
\label{alg:NavigateDiff}
\begin{algorithmic}[1]
\REQUIRE Predictor $\mathcal{G}$, Fusion Policy $\mathcal{\pi}$, time limit $T$ , future frame generation interval $k$, goal image $x_g$, instruction $y$

\STATE $t \leftarrow 0$

\WHILE{$t < T$}
    \STATE $x_t^* \leftarrow \mathcal{G}(x_t,x_g,y)$
    \FOR{$i = 1$ to $k$}
        \STATE $a_t \leftarrow \mathcal{\pi}(x_t^*,x_t,x_g)$
        \STATE Execute $a_t$
        \STATE $x_{t+1} \leftarrow $ Environment
        \STATE $t \leftarrow t+1$
    \ENDFOR
\ENDWHILE

\end{algorithmic}
\end{algorithm}

\begin{figure*}[t]
    \centering
    \includegraphics[width=\textwidth]{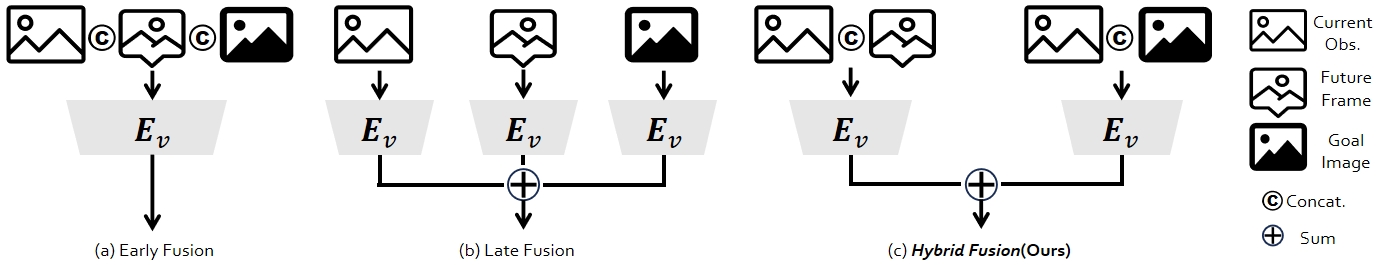}
    \caption{Comparison of different image fusion strategies (a)~Early Fusion, (b)~Late Fusion, and (c)~\textbf{Hybrid Fusion(Ours)}.}
    \label{fig:fusion}
\end{figure*}

\noindent \textbf{Fusion Strategy Design.}
As shown in Fig.~\ref{fig:fusion}, we designed a Hybrid Fusion approach to fuse image features and compared its performance with Early Fusion and Late Fusion. 
In Early Fusion, the current observation, future frame prediction, and goal image are concatenated along the RGB channels and then passed through a visual encoder for feature extraction. 
While this method can capture pixel-level semantic relationships among the three images, it struggles to effectively associate the logical relationships among them. 
In contrast, Late Fusion processes the three images separately through the visual encoder and then fuses them at the feature level, but this approach fails to capture pixel-level semantic correlations, leading to suboptimal performance.

Our proposed Hybrid Fusion takes a different approach: one branch concatenates the current observation and future frame prediction along the RGB channels, while another branch concatenates the current observation and goal image. 
This not only establishes semantic associations at the pixel level but also separates local information (from the current observation to the future frame prediction) and global information (from the current observation to the goal image) in the temporal dimension, resulting in superior performance. 

\begin{table}[t]
\centering
\caption{Quantitative comparison (FID$\downarrow$, PSNR$\uparrow$, LPIPS$\downarrow$). IP2P has also been fine-tuned using the navigation training data.}
\vspace{-3mm}
\begin{tabular}{@{}lccc@{}}
\toprule
Setting            & FID$\downarrow$ & PSNR$\uparrow$    & LPIPS$\downarrow$   \\ 
\midrule
IP2P\cite{brooks2023instructpix2pix}        & 26.59          & 14.59            & 42.25   \\
\textbf{Predictor(Ours)}    & \textbf{25.93}          & \textbf{14.73}            & \textbf{40.82}   \\
\bottomrule
\end{tabular}
\label{tab:Quantitative_result}
\vspace{-3mm}
\end{table}

\begin{figure}[t]
  \centering 
  \includegraphics[width=0.45\textwidth]{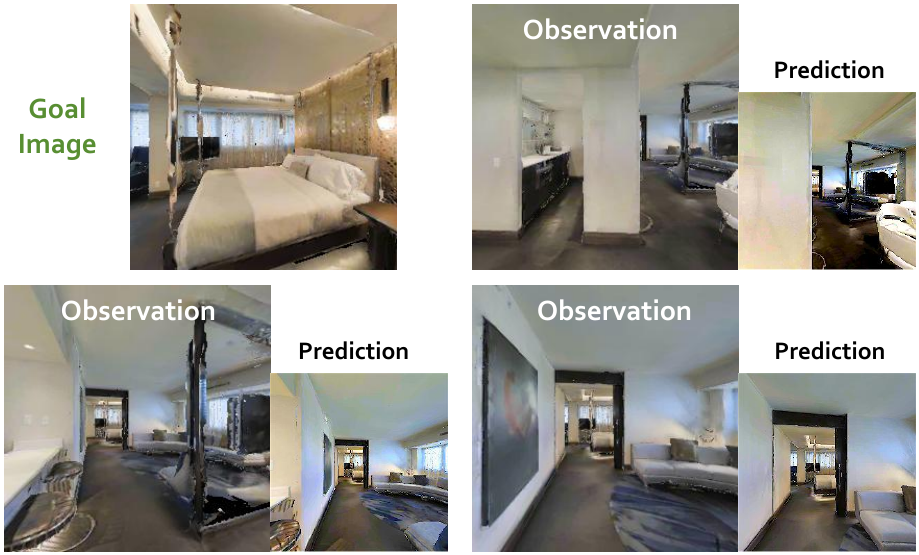}
  \vspace{-3mm}
  \caption{Example rollouts in Gibson dataset.}
  \label{fig:visualization}
\vspace{-7mm}
\end{figure}

\section{Experiments}
\label{sec_experiments}

\subsection{Predictor}
\label{exp_ffp}
\textbf{Dataset.}We used the dataset format constructed in Sec.~\ref{sec_data} and collected video sequences from all image navigation tasks in the GIBSON dataset's training set for training. The interval hyperparameter was set to $k=5$ during the training process.

\textbf{Training Process.} The training process of the Predictor consists of two key stages. First,  InstructPix2Pix~\cite{brooks2023instructpix2pix} is used to pre-train the diffusion model weights in the navigation environment. Next, the Predictor is optimized in an end-to-end fashion. The diffusion model's initial weights in the Predictor are taken directly from those pre-trained in the first stage.

\begin{table}[t]
\centering
\footnotesize
\caption{Comparison with state-of-the-art methods on Gibson.}
\vspace{-3mm}
\begin{tabular}{@{}lccccccc@{}}
\toprule
Method                  & Train Dataset        & Sensor(s)  & SPL           & SR                        \\ 
\midrule
NTS~\cite{chaplot2020neural}          & Full         & RGBD+Pose  & 43.0\%        & 63.0\%                    \\
Act-Neur-SLAM~\cite{chaplot2020neural}   & Full    & RGB+Pose      & 23.0\%        & 35.0\%    \\
SPTM~\cite{savinov2018semi}        & Full         & RGB+Pose   & 27.0\%        & 51.0\%                    \\
\midrule
ZER~\cite{al2022zero}          & Full         & RGB        & 21.6\%        & 29.2\%                    \\
ZSON~\cite{majumdar2022zson}        & Full        & RGB        & 28.0\%        & 36.9\%                    \\
OVRL~\cite{yadav2023offline}        & Full        & RGB        & 27.0\%        & 54.2\%                    \\
OVRL-V2~\cite{yadav2023ovrl}  & Full        & RGB        & 58.7\%        & 82.0\%                    \\
FGPrompt                & Full         & RGB        & 66.5\%        & 90.4\%                    \\
\textbf{NavigateDiff(Ours)} & Full     & RGB        & 64.8\%           & \textbf{91.0\%}                       \\
\midrule
FGPrompt~\cite{sun2024fgprompt}                & 1/4          & RGB        & 48.5\%        & 77.9\%                   \\
\textbf{NavigateDiff(Ours)} & 1/4     & RGB        & \textbf{52.1\%}           & \textbf{81.2\%}     \\
\midrule
FGPrompt~\cite{sun2024fgprompt}                & 1/8          & RGB        & 43.4\%        & 68.1\%                   \\
\textbf{NavigateDiff(Ours)} & 1/8     & RGB        & \textbf{46.4\%}           & \textbf{71.1\%}     \\

\bottomrule
\end{tabular}
\vspace{-5mm}
\label{tab:sota-gibson}
\end{table}

\textbf{Evaluation.} We implement three image-level metrics to evaluate the Predictor's generation ability. (1) Fr\'echet Inception Distance (FID)~\cite{heusel2017gans}, (2) Peak Signal-to-Noise Ratio (PSNR), (3)Learned Perceptual Image Patch Similarity (LPIPS)~\cite{zhang2018unreasonable}. We measure the similarity between the generated future frame and ground truth. In terms of image-level metrics in Tab.~\ref{tab:Quantitative_result}, our Predictor outperforms IP2P by a large margin (0.66, 0.14, and 1.43 respectively) in all three metrics on the Gibson dataset.
In Fig.~\ref{fig:visualization}, we also visualize predicted future frame sequences and trajectory rollouts in the Gibson dataset. We observe that generating future frames one by one could efficiently guide the PolicyNet in action generation.

\subsection{Simulation Experiments}
\textbf{Dataset}
For image-goal navigation, we employ the Habitat simulator and train our agent on the Gibson dataset, using 72 training scenes and 14 testing scenes under the standard configuration. We train the agent for 500M steps, following the rules as outlined in FGPrompt~\cite{sun2024fgprompt}. Results are reported across multiple datasets to enable direct comparison with previous works. For the Gibson dataset, we validate our agent on split A generated by~\cite{mezghan2022memory}. For the MP3D datasets, we use test episodes collected by~\cite{al2022zero}  and the instance image navigation dataset released by~\cite{krantz2022instance}. 

\textbf{Setting}
We report the Success Rate (SR) and Success weighted by Path Length (SPL)~\cite{anderson2018evaluation}, which accounts for the efficiency of the navigation path. An episode is deemed successful if the agent stops within a 1.0m Euclidean distance from the goal. The maximum number of steps per episode is set to 500 by default.
In agent configuration, We follow the setup outlined in prior works~\cite{al2022zero}~\cite{majumdar2022zson}~\cite{yadav2023offline}~\cite{sun2024fgprompt} to initialize an agent equipped with RGB cameras, featuring a 128 × 128 resolution and a 90° field of view (FOV). The agent's action space includes four discrete actions—MOVE\_FORWARD, TURN\_LEFT, TURN\_RIGHT, and STOP—with minimum rotation and forward movement units set to 30° and 0.25m, respectively.

\textbf{Result}
In Tab.~\ref{tab:sota-gibson}, we present a detailed comparison between our model and several state-of-the-art approaches across various metrics.The results highlight the superior performance of our model, particularly in challenging navigation scenarios. To further evaluate the generalization capability of our approach, we conducted additional experiments using a smaller dataset that poses a greater challenge in terms of data availability. Despite the reduced dataset size, our model not only maintained its performance but also outperformed the baseline models. This demonstrates the robustness and adaptability of our model, suggesting it can effectively generalize to new environments even with limited training data. 

As illustrated in Tab.~\ref{tab:cross-domain}, we test the model on the MP3D dataset as part of a cross-task evaluation.  Our \mname  ~achieves 68.0\% Success Rate (SR) and 41.1\% Success weighted by Path Length (SPL) by using a smaller training dataset, surpassing both existing methods on the full dataset and the baseline.

\begin{table}[t]
\centering
\footnotesize
\caption{Cross-domain evaluation on MP3D. The agent is trained in Gibson environments and directly transferred to new environments for evaluation.}
\vspace{-3mm}
\begin{tabular}{@{}lccc@{}}
\toprule
Methods        &               Train Dataset                  & SPL         & SR                  \\ 
\midrule
Mem-Aug~\cite{mezghan2022memory}   & Gibson                       & 3.9\%         & 6.9\%               \\
NRNS~\cite{hahn2021no}         & Gibson                        & 5.2\%         & 9.3\%               \\
ZER~\cite{al2022zero}           & Gibson                         & 10.8\%        & 14.6\%            \\
\midrule
FGPrompt~\cite{sun2024fgprompt}      & 1/4 Gibson                         &37.1\%             & 65.7\%\\ 
\textbf{NavigateDiff (Ours)}            & 1/4 Gibson    & \textbf{41.1\%}   & \textbf{68.0\%}\\ 
\midrule
FGPrompt~\cite{sun2024fgprompt}             & 1/8 Gibson                         &31.8\%             & 55.0\%\\ 
\textbf{NavigateDiff (Ours)}            & 1/8 Gibson    &\textbf{34.9\%}    & \textbf{57.7\%}\\ 
\bottomrule
\end{tabular}
\label{tab:cross-domain}
\end{table}

\begin{table}[t]
\centering
\scriptsize
\caption{Comparing in three diverse real-world environments scenes. (Office, Parking Lot and Corridor).}
\vspace{-3mm}
\begin{tabular}{@{}lccccccc@{}}
\toprule
\multirow{2}{*}{Methods}  & \multicolumn{2}{c}{Office} & \multicolumn{2}{c}{Parking Lot} & \multicolumn{2}{c}{Corridor} \\ 
\cmidrule(l){2-7} 
                     & SPL         & SR         & SPL         & SR       & SPL         & SR   \\ 
\midrule

FGPrompt~\cite{sun2024fgprompt}            &41.0\%& 50.0\%& 52.3\%& 64.3\% & 54.8\%& 71.4\%\\ 
\textbf{NavigateDiff (Ours)}        & \textbf{41.2\%}& \textbf{57.1\%}& 51.1\%& \textbf{71.4\%} & \textbf{55.6\%}& \textbf{78.6\%}\\ 
\bottomrule
\end{tabular}
\vspace{-2mm}
\label{tab:real-world}
\end{table}

\begin{figure}[t]
  \centering 
  \includegraphics[width=0.45\textwidth]{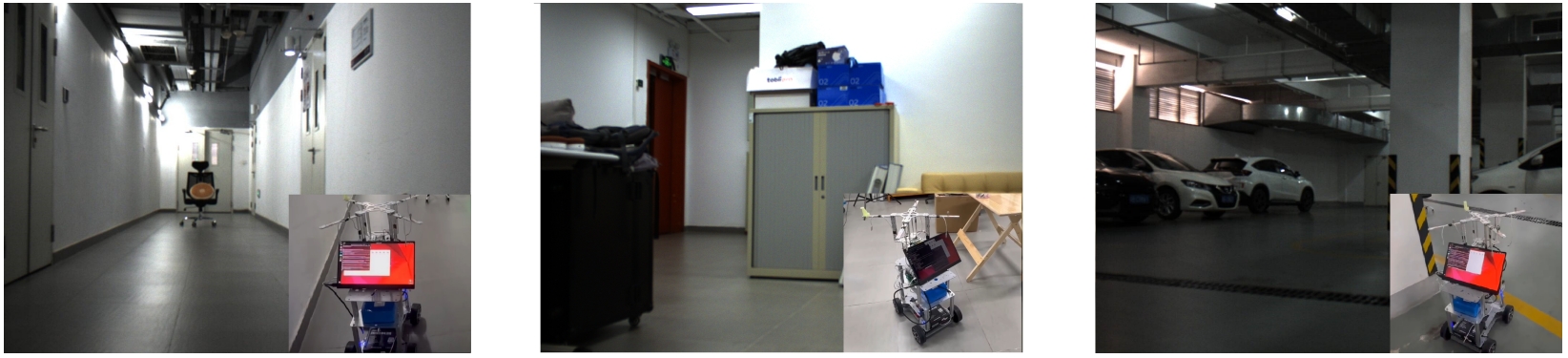}
  \vspace{-2mm}
  \caption{Scenes in Real-world Environment, from the left, in the order of Corridor, Office and Parking Lot}
  \label{fig:real-world scenes}
\end{figure}

\begin{table}[t]
\centering
\small
\caption{Different fusion strategies on Gibson ImageNav task.}
\vspace{-3mm}
\begin{tabular}{@{}lcc@{}}
\toprule
Setting                       & SPL             & SR     \\ 
\midrule
Late Fusion                  & 11.7\%          & 13.7\% \\
Early Fusion                  & 20.5\%          & 40.1\% \\
\textbf{Hybrid Fusion} & \textbf{64.8\%} & \textbf{91.0\%} \\
\bottomrule
\end{tabular}
\label{tab:fusion}
\vspace{-3mm}
\end{table}

\subsection{Real-world Experiments}
\textbf{Setting}
 In our real-world experiments, we focused on indoor environments to evaluate the zero-shot navigation capabilities of our model, \mname. As illustrated in  Fig.~\ref{fig:real-world scenes}, we conducted tests in three types of indoor environments: an office, a parking lot, and a corridor. Each environment represents a unique set of challenges in terms of layout, lighting, and obstacles. 
The office setting is characterized by cluttered spaces, including desks, chairs, and other furniture. 
The indoor parking lot represents a semi-structured environment with clearly defined paths and open spaces but is filled with parked vehicles that act as static obstacles. 
The corridor is a long, narrow space with fewer obstacles but presents challenges in terms of navigation through tight spaces and sharp turns.

\textbf{Result}
As detailed in Tab.~\ref{tab:real-world}, We evaluate the performance of \mname ~in terms of success rate and SPL. The metric across the three real-world scenarios demonstrates that our model consistently surpasses the baseline.

Overall, \mname ~demonstrates strong zero-shot navigation capabilities across all environments. The model's ability to adapt to different layouts and lighting conditions without prior training in those specific environments highlights its robustness in real-world applications.

\subsection{Fusion Strategy Design}
As shown in Tab.~\ref{tab:fusion}, we evaluate different fusion strategies on the Gibson ImageNav task. Our proposed Hybrid Fusion achieves 91.0\% SR and 64.8\% SPL, significantly outperforming both Early Fusion and Late Fusion. These results demonstrate the effectiveness of Hybrid Fusion in integrating future frames into the navigation policy.

\section{Conclusions}
\label{sec_conclusions}
We introduced \mname , a novel approach that leverages logical reasoning and generalization capabilities of pre-trained foundation models to enhance zero-shot navigation by predicting future observations for robust action generation. By integrating temporal information and using a Hybrid Fusion framework to guide policy decisions, our approach significantly improves navigation performance. Extensive experiments demonstrate its efficiency and adaptability in both simulated and real-world environments.

\end{document}